\documentclass[]{hld2023}
\usepackage{multicol}
\usepackage{hyperref}

\newcommand{\Cov}{\mathrm{Cov}}

\hldauthor{%
\Name{Davis Brown\nametag{\thanks{Equal contribution.}
  \addtocounter{footnote}{-1}\addtocounter{Hfootnote}{-1}}} \Email{davis.brown@pnnl.gov}\\
\addr {Pacific Northwest National Laboratory} 
\AND
\Name{Nikhil Vyas\nametag{\footnotemark}} \Email{nikhil@g.harvard.edu}\\
\addr {Harvard University} 
\AND
\Name{Yamini Bansal} \Email{ybansal@google.com}\\
\addr {Google DeepMind} 
}

\title{On Privileged and Convergent Bases in Neural Network Representations}

\begin{document}

\maketitle

\begin{abstract}

In this study, we investigate whether the representations learned by neural networks possess a privileged and convergent basis. Specifically, we examine the significance of feature directions represented by individual neurons. First, we establish that arbitrary rotations of neural representations cannot be inverted (unlike linear networks), indicating that they do not exhibit complete rotational invariance. Subsequently, we explore the possibility of multiple bases achieving identical performance. To do this, we compare the bases of networks trained with the same parameters but with varying random initializations. Our study reveals two findings: (1) Even in wide networks such as WideResNets, neural networks do not converge to a unique basis; (2) Basis correlation increases significantly when a few early layers of the network are frozen identically.

Furthermore, we analyze Linear Mode Connectivity, which has been studied as a measure of basis correlation. Our findings give evidence that while Linear Mode Connectivity improves with increased network width, this improvement is not due to an increase in basis correlation.

\end{abstract}

\section{Introduction}
\label{sec:introduction}

While neural networks are black-box function approximators that are trained end-to-end to optimize a loss objective, their emergent internal layer-wise representations are important objects for both \emph{understanding deep learning} and \emph{direct downstream use}. Internal representations of neural networks can be useful tools for interpretability \cite{olah2020an, cammarata2020curve, bills2023language}, teaching us how neural networks perform the computations they do, as well as for understanding the implicit biases of gradient-based neural network training \cite{ainsworth2023git, model_stitching_bansal}. Moreover, representations are often directly used for downstream tasks that the network was not originally trained for, like in transfer learning or representation learning. Thus, we would like to develop a better understanding of the mathematical properties of neural network representations.

One such property is whether neural networks representations have a \emph{privileged basis} \cite{elhage2022superposition}. That is, are the features represented by each individual neuron significant, or is information stored only at a population level in neurons? This question is important, for instance, in interpretability, where attempts have been made to interpret features represented by individual neurons (such as edge or curve detectors in convolutional networks \cite{cammarata2020curve}). This question is also closely related to that of \emph{invariances} exhibited by neural representations --- what are the set of transformations that can be applied to representations that keep the final network accuracy unchanged? In particular, if the representations are rotation invariant, then an individual neuron does not carry significant information.


To understand this further, consider the simple case of a two-layer neural network without any non-linear activation functions. That is, the function output of the network is $f(x) = W_2W_1x$ with weights $W_2, W_1$ and inputs $x$. Here, the first layer representations are $W_1x$. This representation exhibits rotation invariance - we can rotate the first layer representations by an arbitrary orthonormal matrix $O$ (giving us rotated first layer representations $OW_1x$) but the subsequent layer can invert this rotation and recover the original function $f(x) = W_2O^{-1}OW_1x$. Thus, an individual neuron could represent any arbitrary feature for the same functional outputs in the network.

{\bf Our Contributions:} In our work, we start by showing that the perceived permutation invariance of representations at high width is actually a result of a of kind of noise-averaging --- the correlation between the activities of neurons after accounting for permutations remains \textit{nearly} constant, as we scale the width (Section \ref{sec:sgd-basis}). This shows that while metrics like linear mode connectivity may suggest permutation invariance, the effect disappears when examined at a neuron level. Since this casts some doubt on the presence of a privileged basis, in Section \ref{sec:rand-rot} we ask if \emph{any} basis of neural representations is likely to work equally well. To do so, we consider a random rotation of a layer, and ask if it can be inverted by the later layers with training. We find that this is not the case. Thus, combined these results suggest that while the basis of neural representations matter, there is no one unique basis that is required to achieve the same functional accuracy. Finally, in Section \ref{sec:towards-unique}, we ask what kinds of constraints can be imposed on the network to obtain a unique neural basis consistent across different training runs.

\section{Related Work}
{\bf Convergent learning.} Also referred to as \textit{universality}, convergent learning is the conjecture that different deep learning models learn very similar representations when trained on similar data \cite{DBLP:conf/nips/LiYCLH15, olah2020an}. 
Much of the work in mechanistic interpretability \cite{olah2020an, elhage2021mathematical, olsson2022context, nanda2023progress} has leveraged the universality conjecture to motivate research for toy models, with the hope that the methods and interpretations developed for these more tractable models will scale to larger and more capable models. 
Recently, \cite{chughtai2023toy} examined universality by reverse-engineering a toy transformer model for a group composition task.
Attempts to test for convergent learning include \textit{representation dissimilarity} comparisons, notably neuron alignment \cite{DBLP:conf/nips/LiYCLH15, li2020representation, godfrey2022on} and correlation analysis / centered kernel alignment \cite{morcos2018importance, kornblith2019similarity}.

Model stitching \cite{lenc2014understanding, model_stitching_bansal} extracts features from the early layers of model $f$ and inserts them into the later layers of model $g$ (usually via a learned, low-capacity connecting layer \( \varphi \)). If the representations between these models can be combined such that the resulting `stitched' model, \(g_{>l}\circ \varphi \circ f_{\leq l} \), achieves a low loss on a downstream task, the models are called `stitching connected' for the layer $l$ for that task.

{\bf Linear Mode Connectivity:} It has been conjectured in \cite{EntezariSSN22, ainsworth2023git, jordan2023repair} that, for different models learned by SGD with equal loss, once the permutation symmetries of neural networks are taken into consideration, linear interpolations between them of the form $\theta_\alpha=(1-\alpha) \theta_1+\alpha \theta_2$ for $0<\alpha<1$ have at least constant loss. 

{\bf Privileged Basis:} It is often taken for granted in the interpretability literature that the activation basis is \textit{privileged}, at least in layers with elementwise operations (namely nonlinearities) \cite{erhan2009visualizing, zeiler2014visualizing, zhou2014object,
yosinski2015understanding, Bau2017NetworkDQ}. 
On the other hand, while the residual stream of transformer models has no obvious elementwise operation and is thus not an obviously priveleged basis, \cite{dettmers2022llm} provides evidence for outlier basis-aligned features.
To study this phenomenon, \cite{elhage2023basis} demonstrate that transformers can learn rotationally invariant representations in their residual stream using a similar procedure to what we describe in Section \ref{sec:rand-rot}.
We ask the complementary question of whether layers with nonlinearities can learn in an arbitrary basis.





\section{SGD Basis and Linear Mode Connectivity}\label{sec:sgd-basis}

In this section we explore whether two neural networks with different random initializations converge to the same basis. In other words, are the two neural networks the same up to a permutation of neurons per layer? It is clear that this will hold for an infinite width network since at infinite width there is no `randomness' in initialization. We are interested in answering whether it holds for large yet feasible widths. Recently \cite{EntezariSSN22, ainsworth2023git, jordan2023repair} have studied the weaker claim of if different neural networks are linear mode connected after an appropriate permutation to the neurons per layer. Specifically they study the loss/error barrier when interpolating\footnote{Following~\citet{jordan2023repair}, we reset batch norm statistics in all of the experiments.} between the two networks (after permutation).  We will call this barrier as perm-LMC barrier. In \cite{ainsworth2023git, jordan2023repair} it was found that for networks trained with SGD on CIFAR-10, the perm-LMC barrier become very small (\(<  2\%\) error barrier) for large yet feasible widths. On the other hand, for ImageNet trained models, while the barrier decreases with width, it remains quite large. There are at least two reasons for why perm-LMC barrier could decrease with width:
\begin{enumerate}
    \item The bases of two trained networks becomes closer with larger width.
    \item LMC improves with width even if the two networks do not become closer in their basis with width.
\end{enumerate}

To answer this question, we need a measure of how close two networks are to being permutations of each other. Let $\text{Perm}_n$ be the set of permutations over $n$ elements. For a layer with $n$ neurons and activations $X^1$ and $X^2$ of the two networks, we use $ \max_{p \in \text{Perm}_n} \sum_{i \in [n]} \Cov(X^1_i, X^2_{p(i)}) $ which we call permutation-correlation (perm-corr)\footnote{For all experiments we will report values of perm-corr averaged across all the layers.}. This is also the measure used by~\citet{li2020representation} and~\citet{jordan2023repair} to find an appropriate permutation for calculating the permutation-LMC barrier.


\begin{figure*}[!h]
    \begin{multicols}{3}
    \centering
    \includegraphics[width=0.33\textwidth]{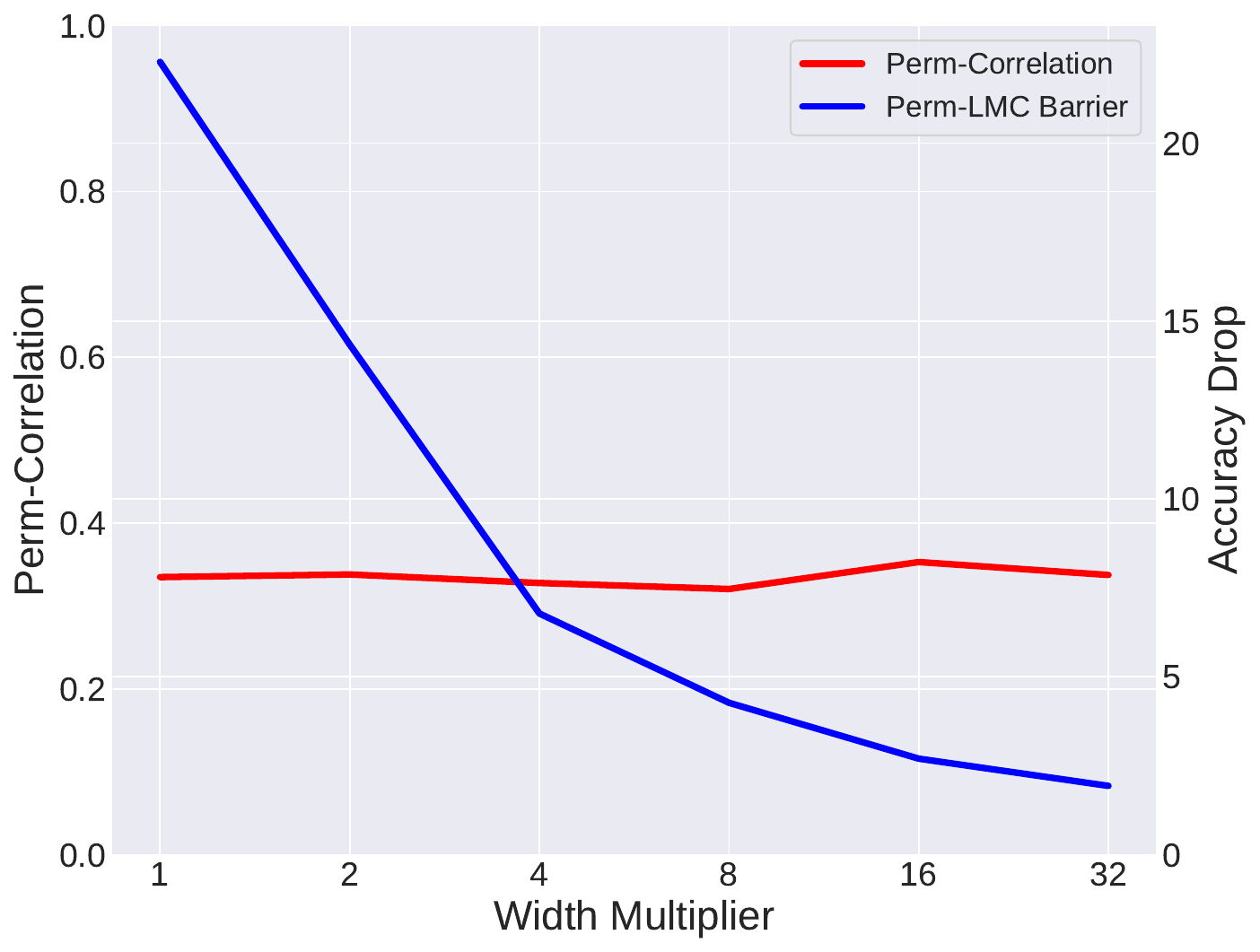} \\
    \includegraphics[width=0.31\textwidth]{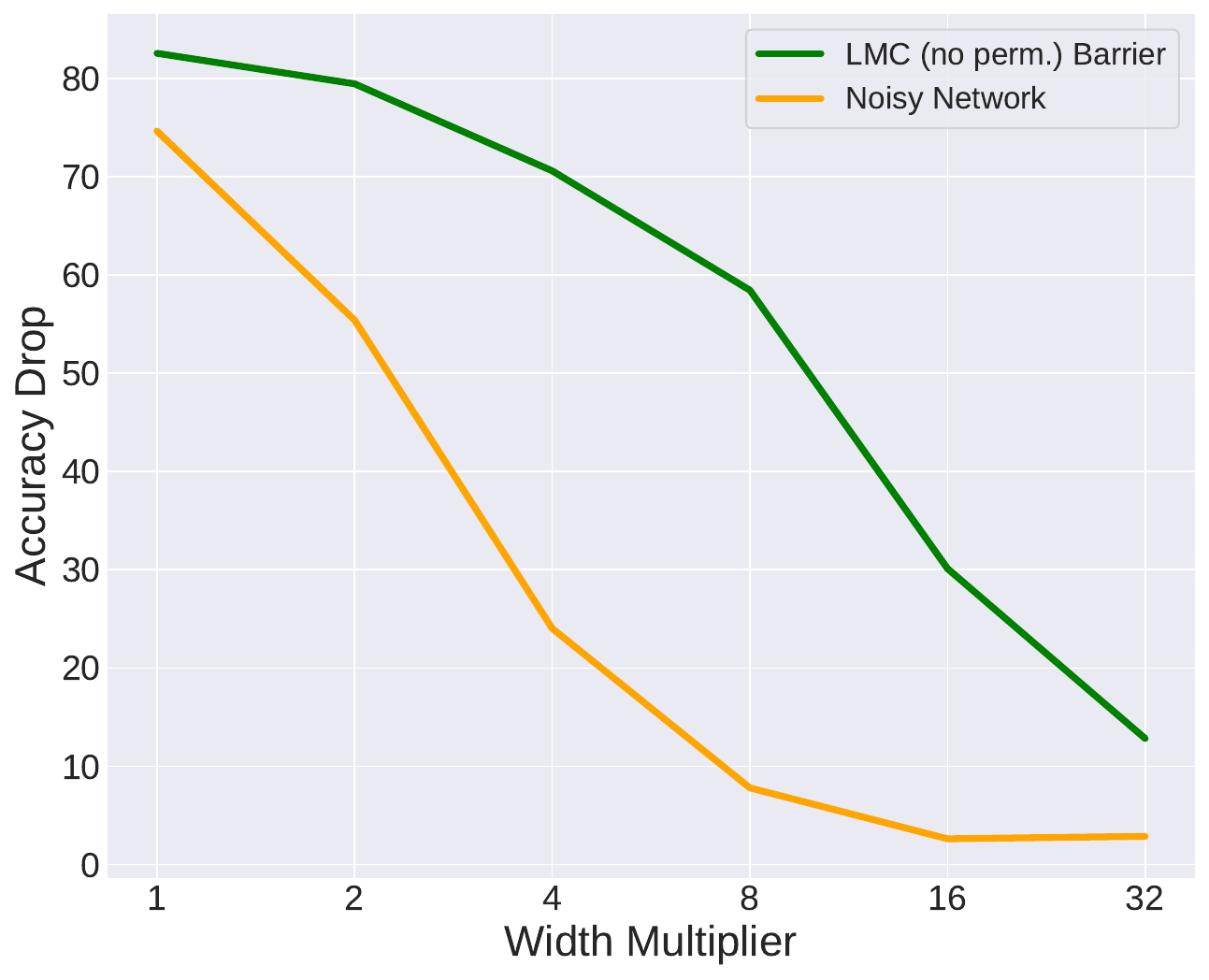} \\
    \includegraphics[width=0.31\textwidth]{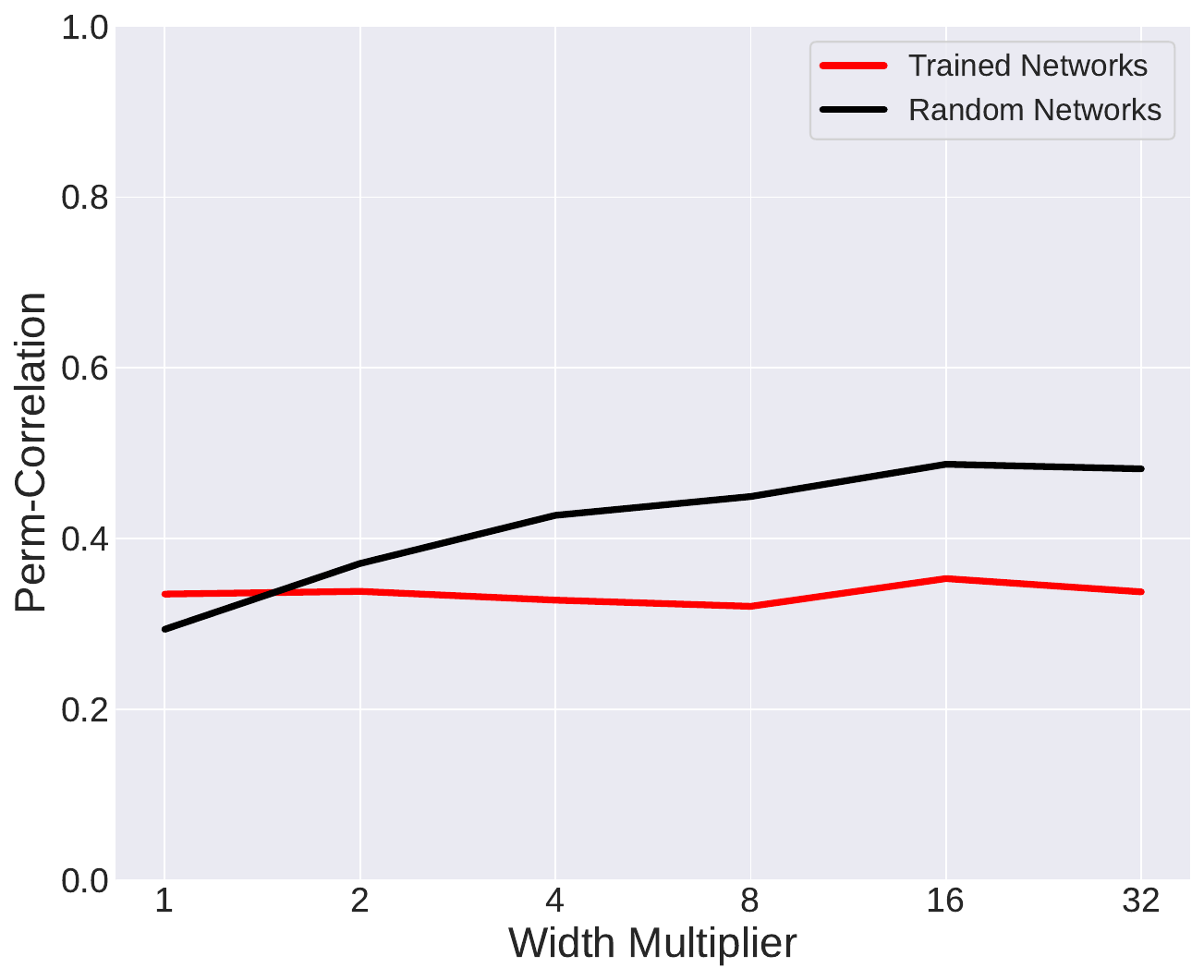} \\
    \end{multicols}
    \caption{Do randomly intialized neural networks converge to the same basis? (\textbf{left}) perm-LMC barrier drops with width but perm-corr is nearly constant.  (\textbf{middle}) LMC barrier (without permutation) also drops with width. (\textbf{right}) perm-corr does not improve through training for most widths.}
    \label{fig:lmc}
\end{figure*}


\paragraph{perm-corr is nearly constant across width: } In Figure~\ref{fig:lmc} (left) we use the exact setup of Jordan et al.~\cite{jordan2023repair} and plot perm-corr and perm-LMC barrier (between two networks trained with independent initializations and batch orderings) as a function of width where we use Resnet20 as the base network and Resnet20-32x as the widest one. perm-corr remains almost constant with width. This suggests that we are very far from widths needed for neurons basis to be close i.e. perm-corr to be near 1. As in prior works we do find that perm-LMC barrier goes down with width. But combined these results suggest that \textbf{perm-LMC barrier goes down with width because linear mode connectivity becomes more feasible at high width rather than due to a better match between the basis of neurons}. We now explore this further.

The previous experiment already verified that basis correlation (operationalized by perm-corr) does not improve with width. We now verify that linear mode connectivity, even without an increase in similarity of basis, improves with width. To do so, we consider the LMC barrier\footnote{\citet{jordan2023repair} did not reset batch norm statistics for this experiment and hence did not observe this improvement with width.} between two networks without permuting the neurons.

In Figure~\ref{fig:lmc} (middle) we find that this quantity (green) improves significantly with width, adding support to our hypothesis. We can now ask why width has this effect. Intuitively, the LMC barrier for two uncorrelated networks (or even networks with some fixed amount of correlation) improves if the networks are robust to noise. This is because being robust to noise means that we can treat the other network as noise and maintain the performance of the first network. Figure~\ref{fig:lmc} (middle) we plot (orange) the performance of a network formed by averaging a trained and a random network (with same weight norms). We find that the accuracy drop (compared to the trained network) of this network also improves with  width adding support to our hypothesis. Together these experiments suggest that that while perm-LMC barrier going down with width is an interesting empirical finding in its own right, it overestimates the similarity of neurons across different networks and in particular is driven by other factors such as robustness to noise to a large extent.

\subsection{Change in Permutation-Correlation due to Training}
Another approach to think about the arising of a unique basis is to see how to training process affects perm-corr i.e., does perm-corr improve due to training? In Figure~\ref{fig:lmc} (right) we plot perm-corr for a trained and random network and find that two \textit{random} networks have higher perm-corr then for two trained network, for all but the narrowest width networks. This suggests that there is a large variance in the basis of neurons that are found in a trained network. This argues against convergent learning in neural networks from the perspective of the basis of neurons.

\section{Can Networks Learn Rotationally Invariant Representations?}\label{sec:rand-rot}
In the last section we saw the neuron basis for networks trained from independent initialization are not aligned (compared to the baseline of two randomly initialized networks), i.e. there is no unique basis. This raises the following natural question: Do all bases lead to good representations?  Specifically, in this section,  we ask if the representations learnt by neural networks can be made rotation invariant. We build on the linear example considered in the introduction, where any orthonormal transformation of the representation can be inverted by the following layer successfully.

To test this in non-linear networks trained with gradient descent, we take a Myrtle-CNN trained with SGD on CIFAR-10 trained to 92\% accuracy. Let's say the network is $f(x) = \sigma(A_l(\sigma(A_{l-1}(...))))$, where $A_l$ denotes the pre-activation of layer $l$ and $\sigma$ denotes the non-linearity. Then, we perform the following procedure: We sample a random orthonormal matrix $O_1$ and multiply it with the \emph{pre-activations} of the first layer $O_1A_1$. Then, we freeze this rotated layer, and retrain all the remaining layers of the network on the same training dataset. Next, we apply a random orthonormal matrix to the second layer $O_2A_2$, freeze the first two layers and retrain all layers $l>2$. We repeat this procedure successively for all the layers in the network. Thus, $f_{l, \text{rotated}}(x) = \sigma(A_l(\sigma(O_{l-1}A_{l-1}(\sigma(O_{l-2}A_{l-2}(...))))$.

\begin{figure*}[!h]
    \begin{multicols}{2}
    \centering
    \includegraphics[width=0.4\textwidth]{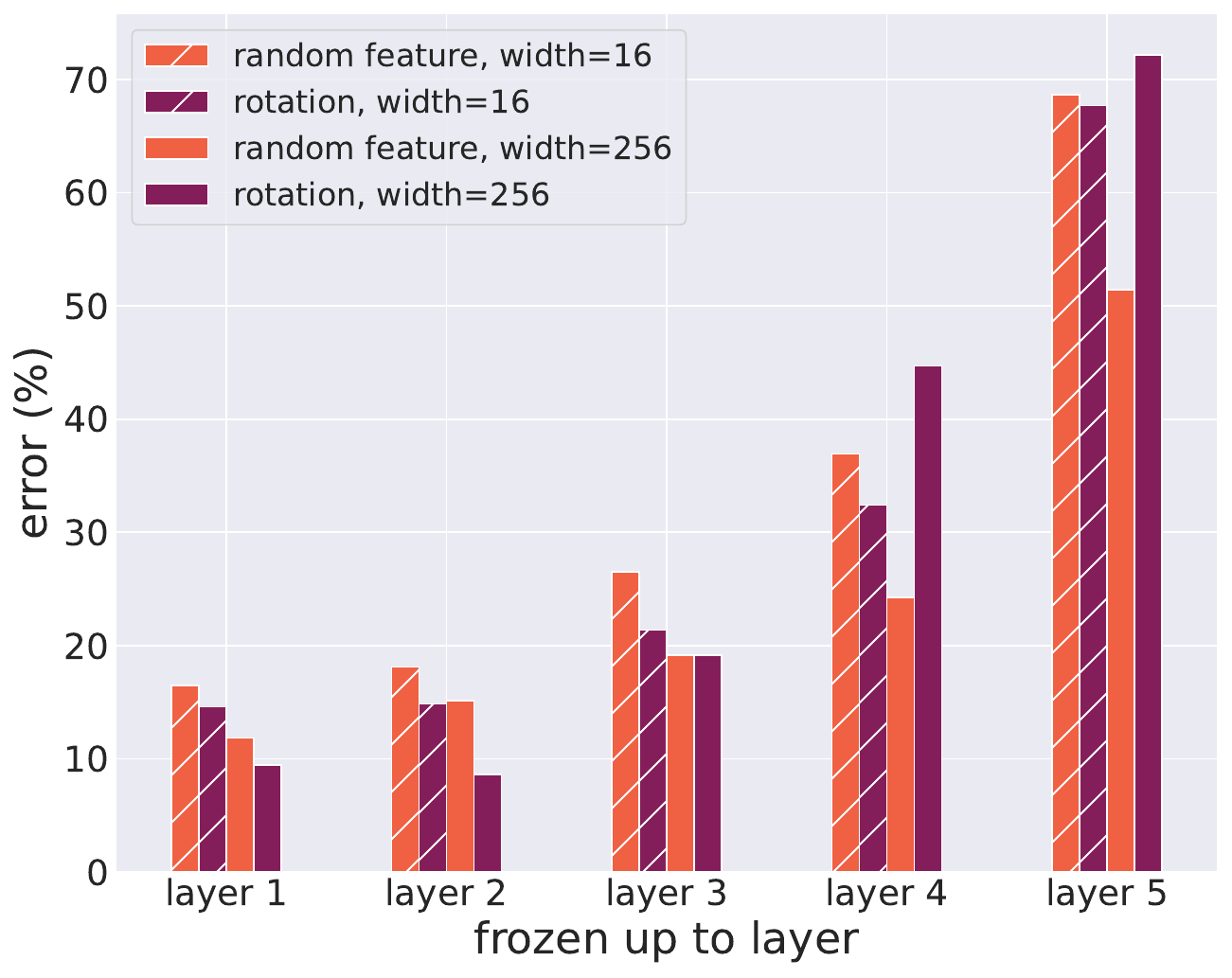}
    \label{fig:random-rot-full}   \\
    \includegraphics[width=0.4\textwidth]{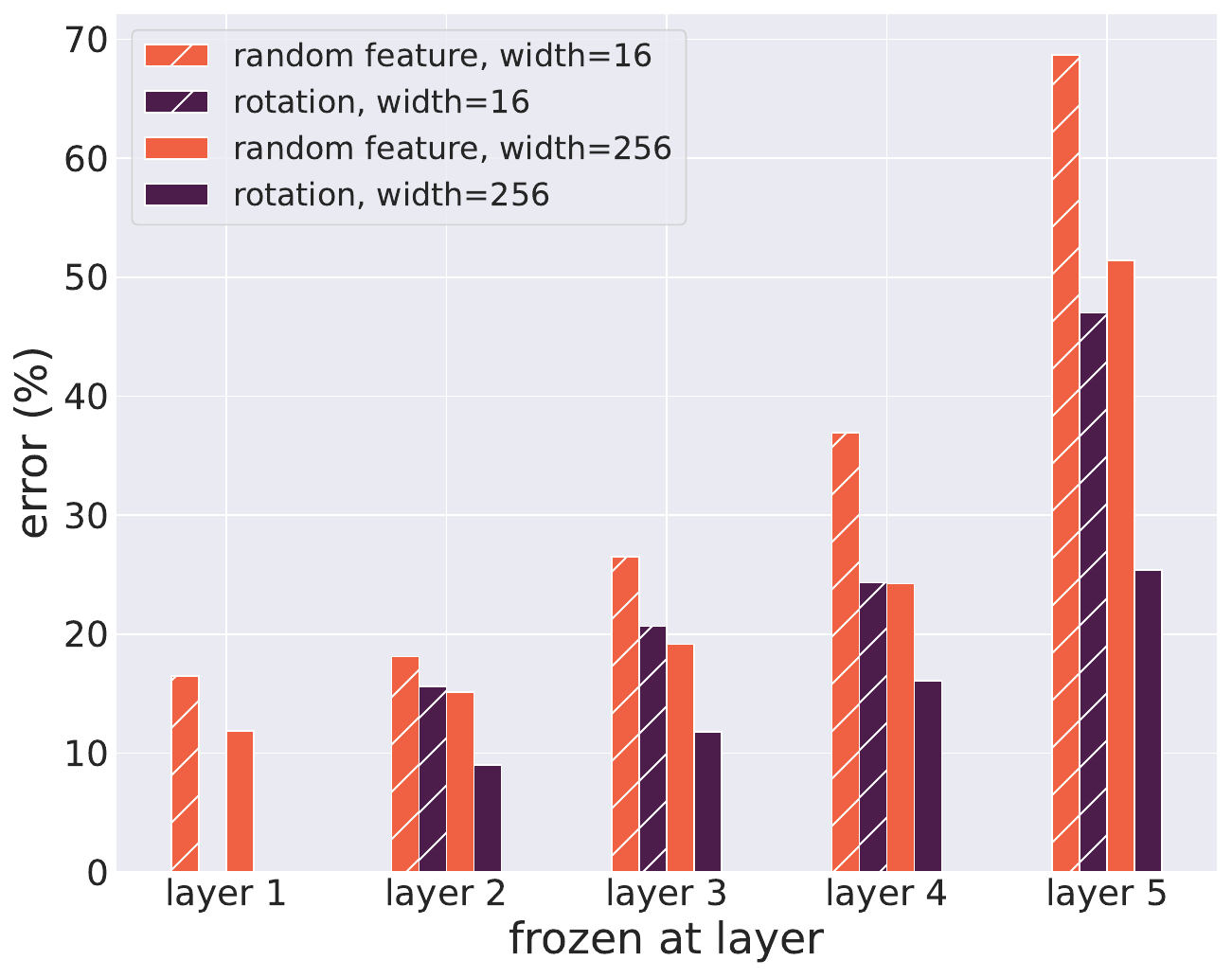}
    \label{fig:random-rot-single}
    \end{multicols}
  \caption{\textit{Random feature} vs \textit{recovering random rotation} performance for Myrtle CNN models trained on CIFAR-10. {(\textbf{left})}  successive freezing, rotating, and retraining; {(\textbf{right})} single layer freezing and rotating.}
  \label{fig:random-rot}
\end{figure*}

Figure \ref{fig:random-rot-full} shows the resulting error when we freeze up to $l$ layers. We find that retraining \emph{cannot} invert this random rotation and that the network accuracy degrades significantly. We also observe that the error gets worse for later layers in the network. For comparison, we also plot the error of a network with random weights (with the same distribution as the initialization) for the first $l$ layers and show that the increase in error from performing random rotations is similar to \textit{random features}-- not training the layers at all! For reference, we repeat the above procedure but we only freeze one layer at a time (Figure \ref{fig:random-rot-single}). We also observe a significant increase in the error of the network, which gets worse as we go deeper into the network. 
Our findings highlight that the basis of the network are not in fact rotation invariant. This suggests that the directions represented by individual neurons are in fact signifcant, and we cannot use an arbitrary basis for the network.


\section{The residual stream might enable convergent bases}\label{sec:towards-unique}

\begin{figure*}[!h]
    \centering
    \begin{multicols}{3}
    {\includegraphics[width=0.31\textwidth]{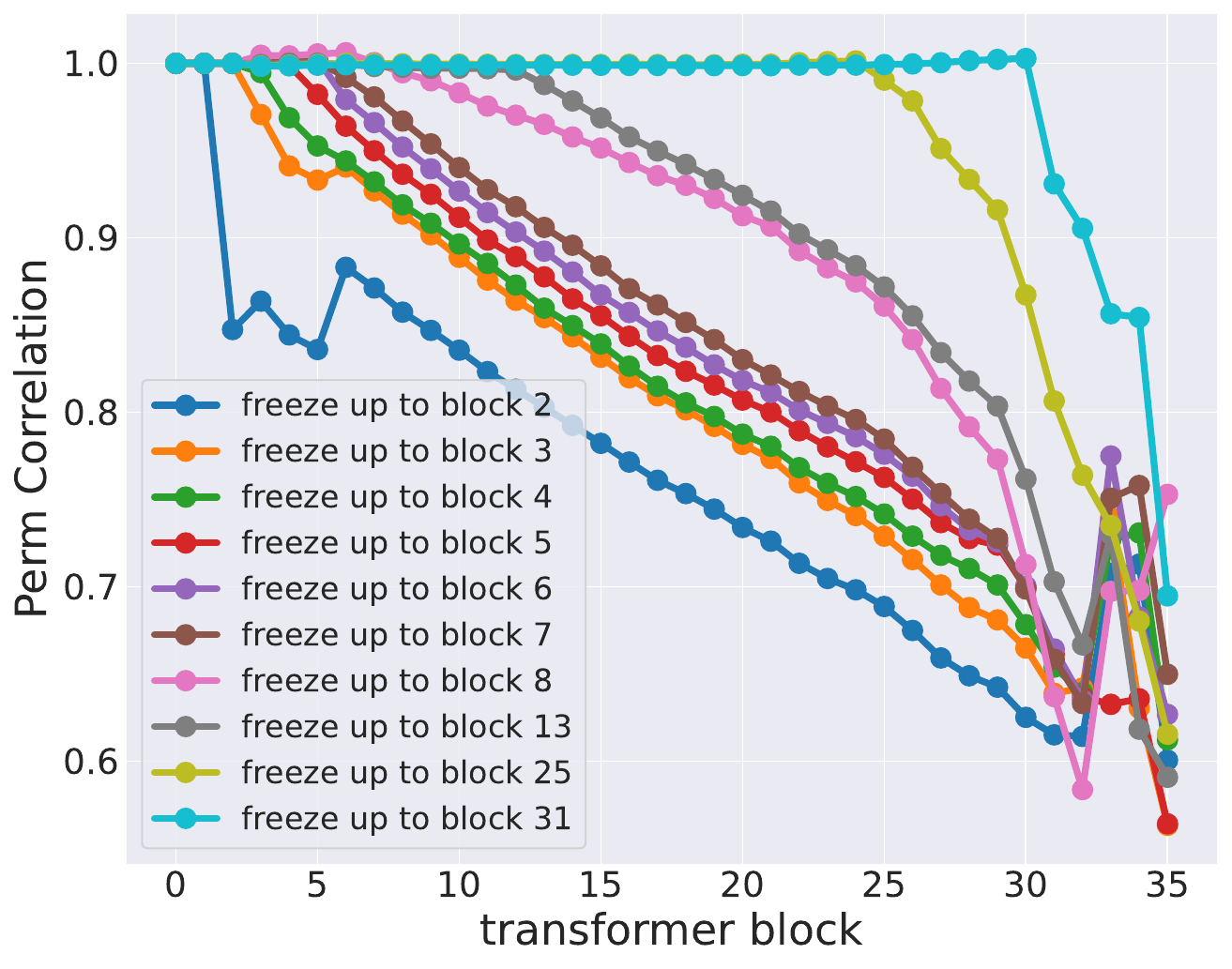}} \\
    {\includegraphics[width=0.31\textwidth]{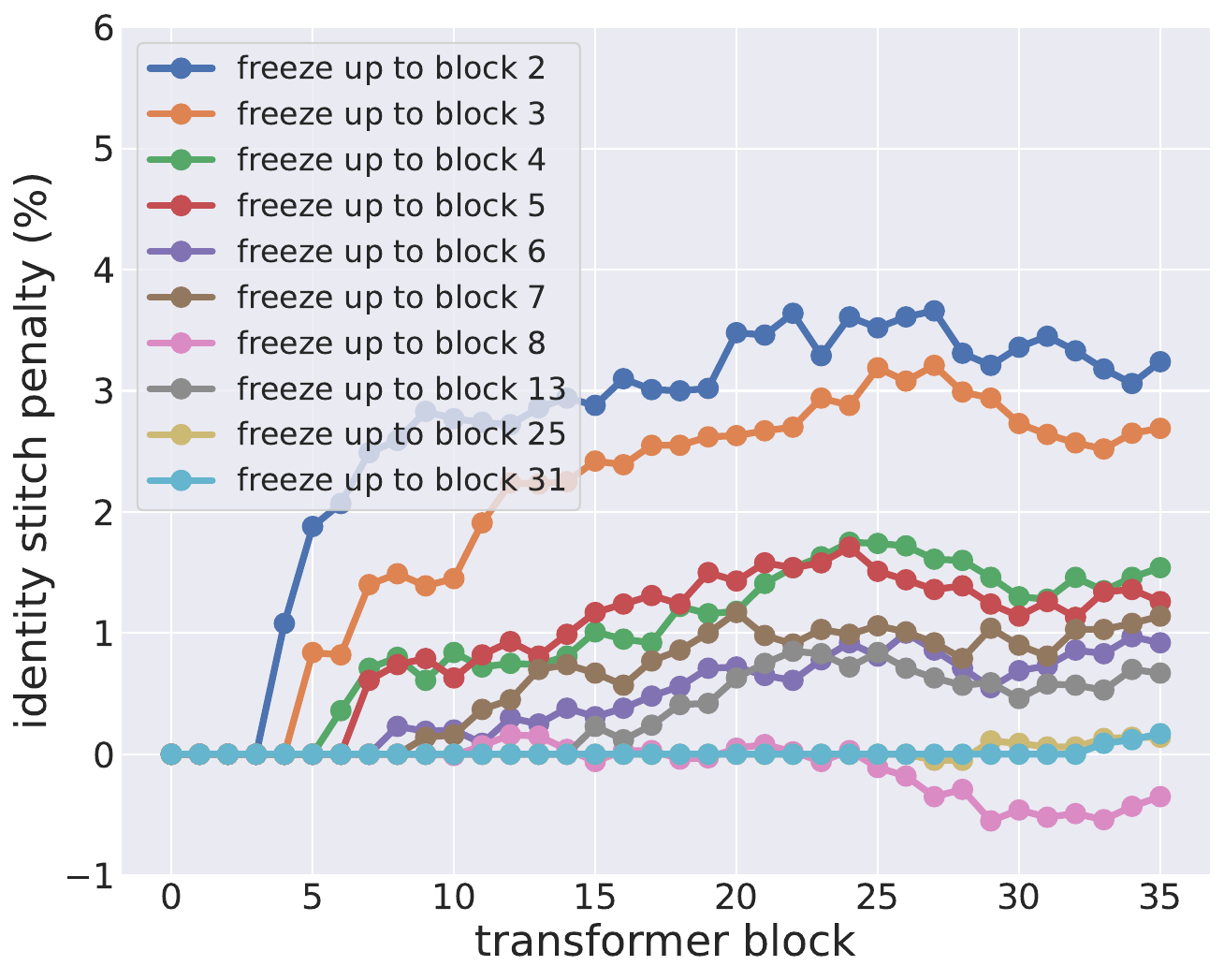}} \\
    {\includegraphics[width=0.31\textwidth]{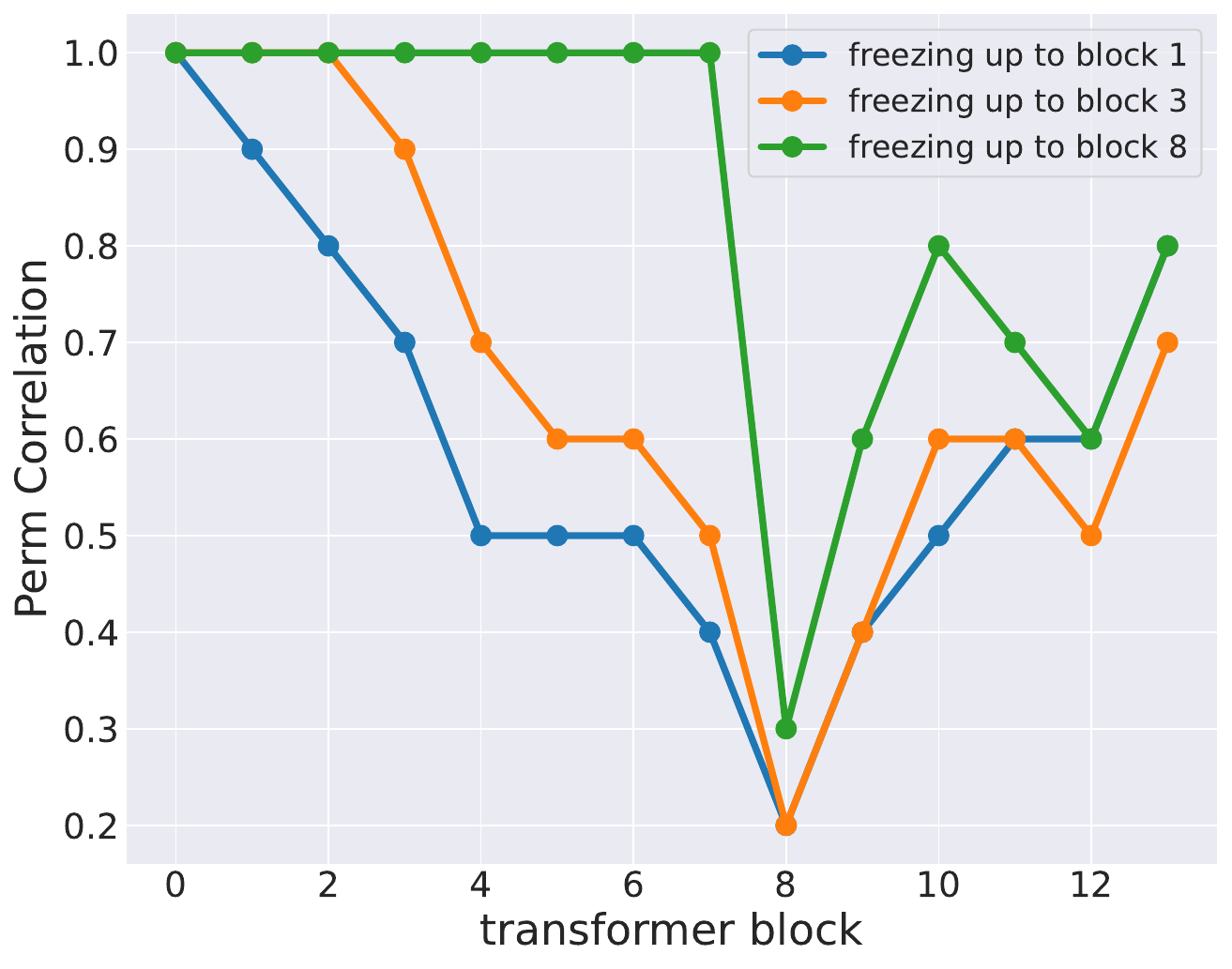}}
    \end{multicols}
    \caption{Enforcing a consistent basis by freezing $l$ early layers. (\textbf{left}) perm-corr for ConvNeXt models trained on CIFAR-10 (\textbf{middle}) identity stitching CIFAR-10 ConvNeXt models (\textbf{right}) perm-corr for ViT models trained on ImageNet}
    \label{fig:res-stream-convergence}
\end{figure*}


Finally, we ask if there are other empirically common choices that may enable the basis to be fixed in certain ways. We highlight the candidate of a \emph{residual stream}. In transformer networks, for instance, there is a largely linear residual connection that runs from the input (after the embedding layer) to the remainder of the network \cite{elhage2021mathematical}. If the residual stream is indeed important to the computations performed by the network, it may impose a basis on the network that roughly tries to align with the input embedding. For models that share the same tokenizer or embedding layer, this might suggest that we can combine networks with residual streams \emph{as is} --- without even having to compute permutations to match the neurons to perform symmetry correction \cite{ainsworth2023git}.

However, prior work \cite{model_stitching_bansal} has shown that the early layers of a vision network can be replaced with random features without significant loss in performance. 
We see in Figure \ref{fig:random-rot} that the first few layers of the network are relatively more resistant to random rotations. 
This suggests that even if the residual connections were providing a force for the basis to align, the residual connections would align to the basis computed by these random transformations of the early layers. 

Thus, we experiment with freezing early blocks of the network. First, we train a model $A$ with $n$ layers. Then, we train a new model $B$ with the first $l$ frozen layers of $A$, training the $n-l$ top-layers from scratch with a new random initialization. The perm-corr is computed between layers, Figure \ref{fig:res-stream-convergence} shows our results for a ConvNeXt \cite{liu2022convnet} model trained on CIFAR-10 (left) and a Vision Transformer (ViT) \cite{dosovitskiy2020image} trained on ImageNet (right).
We also measure the identity stitching penalty, where we evaluate the performance of pairs models stitched with the \emph{identity function} \(\varphi = \mathrm{id}\), i.e. models of the form  \( g_{>l}\circ  f_{\leq l}\).

We find that freezing only 2 blocks is sufficient to have a significantly higher perm-corr (likewise, lower error for identity stitching Figure \ref{fig:res-stream-convergence} (middle)) in ConvNeXt models and ViTs. This convergence phenomenon is considered across different model widths and for different residual stream structures in Appendix \ref{app:towards-unique}.

Together, these results suggest a relatively cheap procedure for \textit{fixing the basis} when training different neural networks, an (as noted) desirable property for interpretability research and also relevant for modular and distributed neural network training \cite{konečný2016federated, raffelOS}. For example, a related phenomenon for networks fine-tuned from the same base model was used in \cite{wortsmanModelSoupsAveraging2022} to create a combined model more accurate than its constituent models.


\section{Discussion and conclusion}
We find that, in some important respects, linear mode connectivity overstates the similarity of neurons across runs. Our results suggest that while the LMC barrier improves with network width, it can in part be explained by factors beyond similarity. On the other hand, we find strong evidence that for neural networks, only certain bases (namely, those that are not rotation invariant) lead to good representations. Finally, we provide a straightforward procedure to enable a convergent basis; this is desirable for both interpretability and modular training.


\section*{Acknowledgements}
NV is supported by a Simons Investigator Fellowship, NSF grant DMS-2134157, DARPA grant W911NF2010021, and DOE grant DE-SC0022199.




\bibliography{ref}

\begin{thebibliography}{30}
\providecommand{\natexlab}[1]{#1}
\providecommand{\url}[1]{\texttt{#1}}
\expandafter\ifx\csname urlstyle\endcsname\relax
  \providecommand{\doi}[1]{doi: #1}\else
  \providecommand{\doi}{doi: \begingroup \urlstyle{rm}\Url}\fi

\bibitem[Ainsworth et~al.(2023)Ainsworth, Hayase, and
  Srinivasa]{ainsworth2023git}
Samuel Ainsworth, Jonathan Hayase, and Siddhartha Srinivasa.
\newblock Git re-basin: Merging models modulo permutation symmetries.
\newblock In \emph{International Conference on Learning Representations}, 2023.
\newblock URL \url{https://openreview.net/forum?id=CQsmMYmlP5T}.

\bibitem[Bansal et~al.(2021)Bansal, Nakkiran, and
  Barak]{model_stitching_bansal}
Yamini Bansal, Preetum Nakkiran, and Boaz Barak.
\newblock Revisiting model stitching to compare neural representations.
\newblock In M.~Ranzato, A.~Beygelzimer, Y.~Dauphin, P.S. Liang, and J.~Wortman
  Vaughan, editors, \emph{Advances in Neural Information Processing Systems},
  volume~34, pages 225--236. Curran Associates, Inc., 2021.
\newblock URL
  \url{https://proceedings.neurips.cc/paper/2021/file/01ded4259d101feb739b06c399e9cd9c-Paper.pdf}.

\bibitem[Bau et~al.(2017)Bau, Zhou, Khosla, Oliva, and
  Torralba]{Bau2017NetworkDQ}
David Bau, Bolei Zhou, Aditya Khosla, Aude Oliva, and Antonio Torralba.
\newblock Network dissection: Quantifying interpretability of deep visual
  representations.
\newblock \emph{2017 IEEE Conference on Computer Vision and Pattern Recognition
  (CVPR)}, pages 3319--3327, 2017.

\bibitem[Bills et~al.(2023)Bills, Cammarata, Mossing, Tillman, Gao, Goh,
  Sutskever, Leike, Wu, and Saunders]{bills2023language}
Steven Bills, Nick Cammarata, Dan Mossing, Henk Tillman, Leo Gao, Gabriel Goh,
  Ilya Sutskever, Jan Leike, Jeff Wu, and William Saunders.
\newblock Language models can explain neurons in language models.
\newblock
  \url{https://openaipublic.blob.core.windows.net/neuron-explainer/paper/index.html},
  2023.

\bibitem[Cammarata et~al.(2020)Cammarata, Goh, Carter, Schubert, Petrov, and
  Olah]{cammarata2020curve}
Nick Cammarata, Gabriel Goh, Shan Carter, Ludwig Schubert, Michael Petrov, and
  Chris Olah.
\newblock Curve detectors.
\newblock \emph{Distill}, 2020.
\newblock \doi{10.23915/distill.00024.003}.
\newblock https://distill.pub/2020/circuits/curve-detectors.

\bibitem[Chughtai et~al.(2023)Chughtai, Chan, and Nanda]{chughtai2023toy}
B.~Chughtai, Lawrence Chan, and Neel Nanda.
\newblock A toy model of universality: Reverse engineering how networks learn
  group operations.
\newblock \emph{ARXIV.ORG}, 2023.
\newblock \doi{10.48550/arXiv.2302.03025}.

\bibitem[Dettmers et~al.(2022)Dettmers, Lewis, Belkada, and
  Zettlemoyer]{dettmers2022llm}
Tim Dettmers, Mike Lewis, Younes Belkada, and Luke Zettlemoyer.
\newblock Llm. int8 (): 8-bit matrix multiplication for transformers at scale.
\newblock \emph{arXiv preprint arXiv:2208.07339}, 2022.

\bibitem[Dosovitskiy et~al.(2020)Dosovitskiy, Beyer, Kolesnikov, Weissenborn,
  Zhai, Unterthiner, Dehghani, Minderer, Heigold, Gelly, Uszkoreit, and
  Houlsby]{dosovitskiy2020image}
A.~Dosovitskiy, L.~Beyer, Alexander Kolesnikov, Dirk Weissenborn, Xiaohua Zhai,
  Thomas Unterthiner, M.~Dehghani, Matthias Minderer, G.~Heigold, S.~Gelly,
  Jakob Uszkoreit, and N.~Houlsby.
\newblock An image is worth 16x16 words: Transformers for image recognition at
  scale.
\newblock \emph{International Conference On Learning Representations}, 2020.

\bibitem[Elhage et~al.(2021)Elhage, Nanda, Olsson, Henighan, Joseph, Mann,
  Askell, Bai, Chen, Conerly, et~al.]{elhage2021mathematical}
N~Elhage, N~Nanda, C~Olsson, T~Henighan, N~Joseph, B~Mann, A~Askell, Y~Bai,
  A~Chen, T~Conerly, et~al.
\newblock A mathematical framework for transformer circuits.
\newblock \emph{Transformer Circuits Thread}, 2021.

\bibitem[Elhage et~al.(2022)Elhage, Hume, Olsson, Schiefer, Henighan, Kravec,
  Hatfield-Dodds, Lasenby, Drain, Chen, Grosse, McCandlish, Kaplan, Amodei,
  Wattenberg, and Olah]{elhage2022superposition}
Nelson Elhage, Tristan Hume, Catherine Olsson, Nicholas Schiefer, Tom Henighan,
  Shauna Kravec, Zac Hatfield-Dodds, Robert Lasenby, Dawn Drain, Carol Chen,
  Roger Grosse, Sam McCandlish, Jared Kaplan, Dario Amodei, Martin Wattenberg,
  and Christopher Olah.
\newblock Toy models of superposition.
\newblock \emph{Transformer Circuits Thread}, 2022.
\newblock URL \url{https://transformer-circuits.pub/2022/toy_model/index.html}.

\bibitem[Elhage et~al.(2023)Elhage, Lasenby, and Olah]{elhage2023basis}
Nelson Elhage, Robert Lasenby, and Christopher Olah.
\newblock Privileged bases in the transformer residual stream.
\newblock \emph{Transformer Circuits Thread}, 2023.
\newblock URL
  \url{https://transformer-circuits.pub/2023/privileged-basis/index.html}.

\bibitem[Entezari et~al.(2022)Entezari, Sedghi, Saukh, and
  Neyshabur]{EntezariSSN22}
Rahim Entezari, Hanie Sedghi, Olga Saukh, and Behnam Neyshabur.
\newblock The role of permutation invariance in linear mode connectivity of
  neural networks.
\newblock In \emph{The Tenth International Conference on Learning
  Representations, {ICLR} 2022, Virtual Event, April 25-29, 2022}.
  OpenReview.net, 2022.
\newblock URL \url{https://openreview.net/forum?id=dNigytemkL}.

\bibitem[Erhan et~al.(2009)Erhan, Bengio, Courville, and
  Vincent]{erhan2009visualizing}
Dumitru Erhan, Yoshua Bengio, Aaron Courville, and Pascal Vincent.
\newblock Visualizing higher-layer features of a deep network.
\newblock \emph{University of Montreal}, 1341\penalty0 (3):\penalty0 1, 2009.

\bibitem[Godfrey et~al.(2022)Godfrey, Brown, Emerson, and
  Kvinge]{godfrey2022on}
Charles Godfrey, Davis Brown, Tegan Emerson, and Henry Kvinge.
\newblock On the symmetries of deep learning models and their internal
  representations.
\newblock In Alice~H. Oh, Alekh Agarwal, Danielle Belgrave, and Kyunghyun Cho,
  editors, \emph{Advances in Neural Information Processing Systems}, 2022.
\newblock URL \url{https://openreview.net/forum?id=8qugS9JqAxD}.

\bibitem[Jordan et~al.(2023)Jordan, Sedghi, Saukh, Entezari, and
  Neyshabur]{jordan2023repair}
Keller Jordan, Hanie Sedghi, Olga Saukh, Rahim Entezari, and Behnam Neyshabur.
\newblock {REPAIR}: {RE}normalizing permuted activations for interpolation
  repair.
\newblock In \emph{The Eleventh International Conference on Learning
  Representations}, 2023.
\newblock URL \url{https://openreview.net/forum?id=gU5sJ6ZggcX}.

\bibitem[Konečný et~al.(2016)Konečný, McMahan, Yu, Richtárik, Suresh, and
  Bacon]{konečný2016federated}
Jakub Konečný, H.~Brendan McMahan, Felix~X. Yu, Peter Richtárik,
  Ananda~Theertha Suresh, and Dave Bacon.
\newblock Federated learning: Strategies for improving communication
  efficiency.
\newblock \emph{arXiv preprint arXiv: 1610.05492}, 2016.

\bibitem[Kornblith et~al.(2019)Kornblith, Norouzi, Lee, and
  Hinton]{kornblith2019similarity}
Simon Kornblith, Mohammad Norouzi, Honglak Lee, and Geoffrey Hinton.
\newblock Similarity of neural network representations revisited.
\newblock In \emph{International Conference on Machine Learning}, pages
  3519--3529. PMLR, 2019.

\bibitem[Lenc and Vedaldi(2014)]{lenc2014understanding}
Karel Lenc and A.~Vedaldi.
\newblock Understanding image representations by measuring their equivariance
  and equivalence.
\newblock \emph{Computer Vision And Pattern Recognition}, 2014.
\newblock \doi{10.1007/s11263-018-1098-y}.

\bibitem[Li et~al.(2020)Li, Grandvalet, Flamary, Courty, and
  Dou]{li2020representation}
Xuhong Li, Yves Grandvalet, R{\'e}mi Flamary, Nicolas Courty, and Dejing Dou.
\newblock Representation transfer by optimal transport.
\newblock \emph{arXiv preprint arXiv:2007.06737}, 2020.

\bibitem[Li et~al.(2015)Li, Yosinski, Clune, Lipson, and
  Hopcroft]{DBLP:conf/nips/LiYCLH15}
Yixuan Li, Jason Yosinski, Jeff Clune, Hod Lipson, and John~E. Hopcroft.
\newblock Convergent learning: Do different neural networks learn the same
  representations?
\newblock In \emph{Proceedings of the 1st Workshop on Feature Extraction:
  Modern Questions and Challenges, {FE} 2015, co-located with the 29th Annual
  Conference on Neural Information Processing Systems {(NIPS} 2015), Montreal,
  Canada, December 11-12, 2015}, volume~44 of \emph{{JMLR} Workshop and
  Conference Proceedings}, pages 196--212. JMLR.org, 2015.
\newblock URL \url{http://proceedings.mlr.press/v44/li15convergent.html}.

\bibitem[Liu et~al.(2022)Liu, Mao, Wu, Feichtenhofer, Darrell, and
  Xie]{liu2022convnet}
Zhuang Liu, Hanzi Mao, Chao-Yuan Wu, Christoph Feichtenhofer, Trevor Darrell,
  and Saining Xie.
\newblock A convnet for the 2020s.
\newblock \emph{CVPR}, 2022.

\bibitem[Morcos et~al.(2018)Morcos, Barrett, Rabinowitz, and
  Botvinick]{morcos2018importance}
Ari~S. Morcos, David G.~T. Barrett, Neil~C. Rabinowitz, and Matthew Botvinick.
\newblock On the importance of single directions for generalization, 2018.
\newblock URL \url{https://arxiv.org/abs/1803.06959}.

\bibitem[Nanda et~al.(2023)Nanda, Chan, Lieberum, Smith, and
  Steinhardt]{nanda2023progress}
Neel Nanda, Lawrence Chan, Tom Lieberum, Jess Smith, and Jacob Steinhardt.
\newblock Progress measures for grokking via mechanistic interpretability.
\newblock In \emph{The Eleventh International Conference on Learning
  Representations}, 2023.
\newblock URL \url{https://openreview.net/forum?id=9XFSbDPmdW}.

\bibitem[Olah et~al.(2020)Olah, Cammarata, Schubert, Goh, Petrov, and
  Carter]{olah2020an}
Chris Olah, Nick Cammarata, Ludwig Schubert, Gabriel Goh, Michael Petrov, and
  Shan Carter.
\newblock An overview of early vision in inceptionv1.
\newblock \emph{Distill}, 2020.
\newblock \doi{10.23915/distill.00024.002}.
\newblock https://distill.pub/2020/circuits/early-vision.

\bibitem[Olsson et~al.(2022)Olsson, Elhage, Nanda, Joseph, DasSarma, Henighan,
  Mann, Askell, Bai, Chen, et~al.]{olsson2022context}
Catherine Olsson, Nelson Elhage, Neel Nanda, Nicholas Joseph, Nova DasSarma,
  Tom Henighan, Ben Mann, Amanda Askell, Yuntao Bai, Anna Chen, et~al.
\newblock In-context learning and induction heads.
\newblock \emph{arXiv preprint arXiv:2209.11895}, 2022.

\bibitem[Raffel(2023)]{raffelOS}
Colin Raffel.
\newblock Building machine learning models like open source software.
\newblock \emph{Commun. ACM}, 66\penalty0 (2):\penalty0 38–40, jan 2023.
\newblock ISSN 0001-0782.
\newblock \doi{10.1145/3545111}.
\newblock URL \url{https://doi.org/10.1145/3545111}.

\bibitem[Wortsman et~al.(2022)Wortsman, Ilharco, Gadre, Roelofs,
  {Gontijo-Lopes}, Morcos, Namkoong, Farhadi, Carmon, Kornblith, and
  Schmidt]{wortsmanModelSoupsAveraging2022}
Mitchell Wortsman, Gabriel Ilharco, Samir~Yitzhak Gadre, Rebecca Roelofs,
  Raphael {Gontijo-Lopes}, Ari~S. Morcos, Hongseok Namkoong, Ali Farhadi, Yair
  Carmon, Simon Kornblith, and Ludwig Schmidt.
\newblock Model soups: Averaging weights of multiple fine-tuned models improves
  accuracy without increasing inference time, July 2022.

\bibitem[Yosinski et~al.(2015)Yosinski, Clune, Nguyen, Fuchs, and
  Lipson]{yosinski2015understanding}
Jason Yosinski, Jeff Clune, Anh Nguyen, Thomas Fuchs, and Hod Lipson.
\newblock Understanding neural networks through deep visualization.
\newblock In \emph{Deep Learning Workshop, International Conference on Machine
  Learning (ICML)}, 2015.

\bibitem[Zeiler and Fergus(2014)]{zeiler2014visualizing}
Matthew~D Zeiler and Rob Fergus.
\newblock Visualizing and understanding convolutional networks.
\newblock In \emph{European conference on computer vision}, pages 818--833.
  Springer, 2014.

\bibitem[Zhou et~al.(2015)Zhou, Khosla, Àgata Lapedriza, Oliva, and
  Torralba]{zhou2014object}
Bolei Zhou, Aditya Khosla, Àgata Lapedriza, Aude Oliva, and Antonio Torralba.
\newblock Object detectors emerge in deep scene cnns.
\newblock In \emph{ICLR}, 2015.
\newblock URL \url{http://arxiv.org/abs/1412.6856}.

\end{thebibliography}

\newpage
\appendix
\onecolumn




\section{Additional Experiments for Section~\ref{sec:towards-unique}}\label{app:towards-unique}

\begin{figure*}[!h]
    \centering
  {\includegraphics[width=0.48\columnwidth]{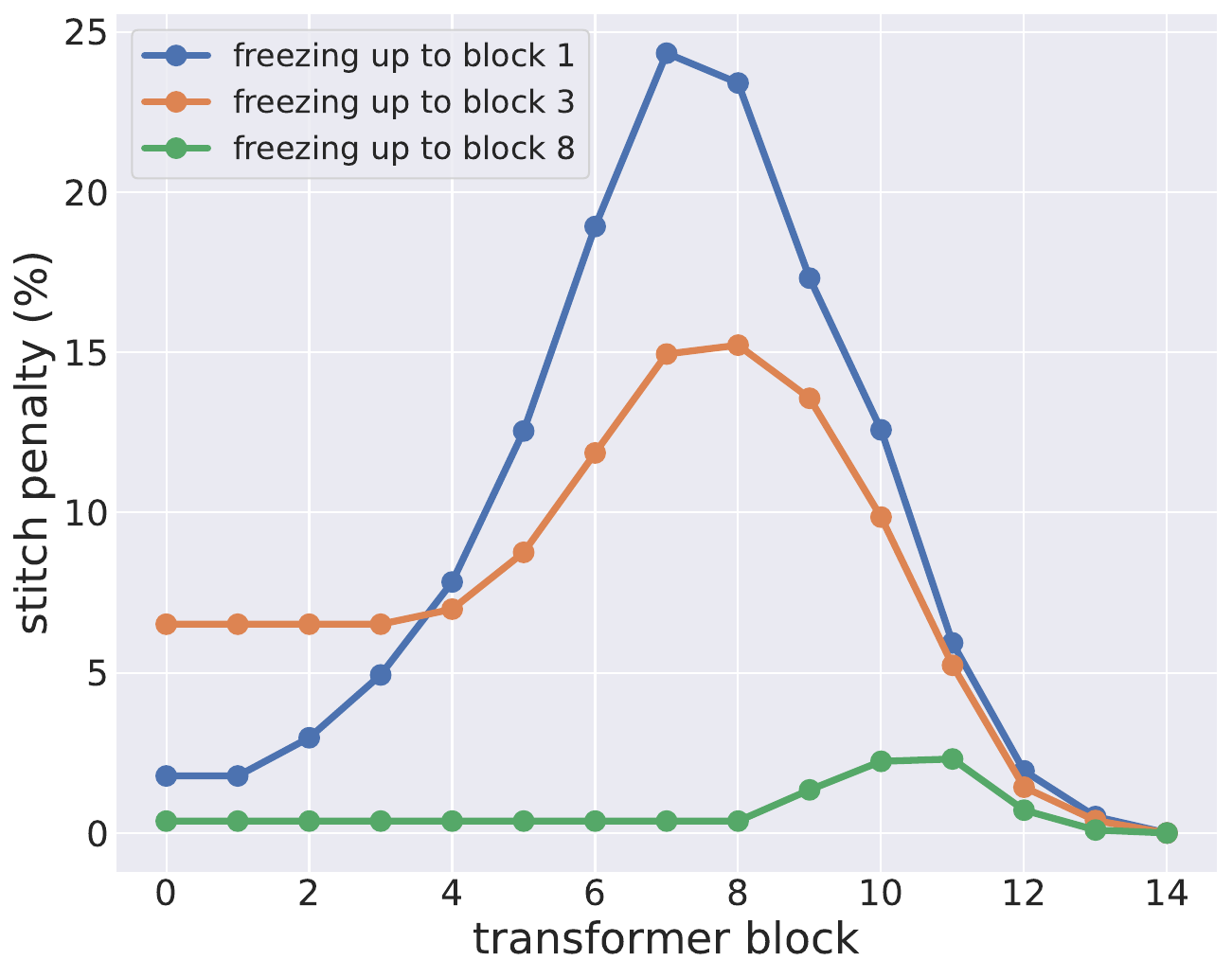}
    }
  \caption{Identity stitching for Vision Transformers trained on ImageNet. }
  \label{fig:random-rot-vit}
\end{figure*}

\begin{figure*}[!h]
    \centering
    \begin{multicols}{3}
    {\includegraphics[width=0.31\textwidth]{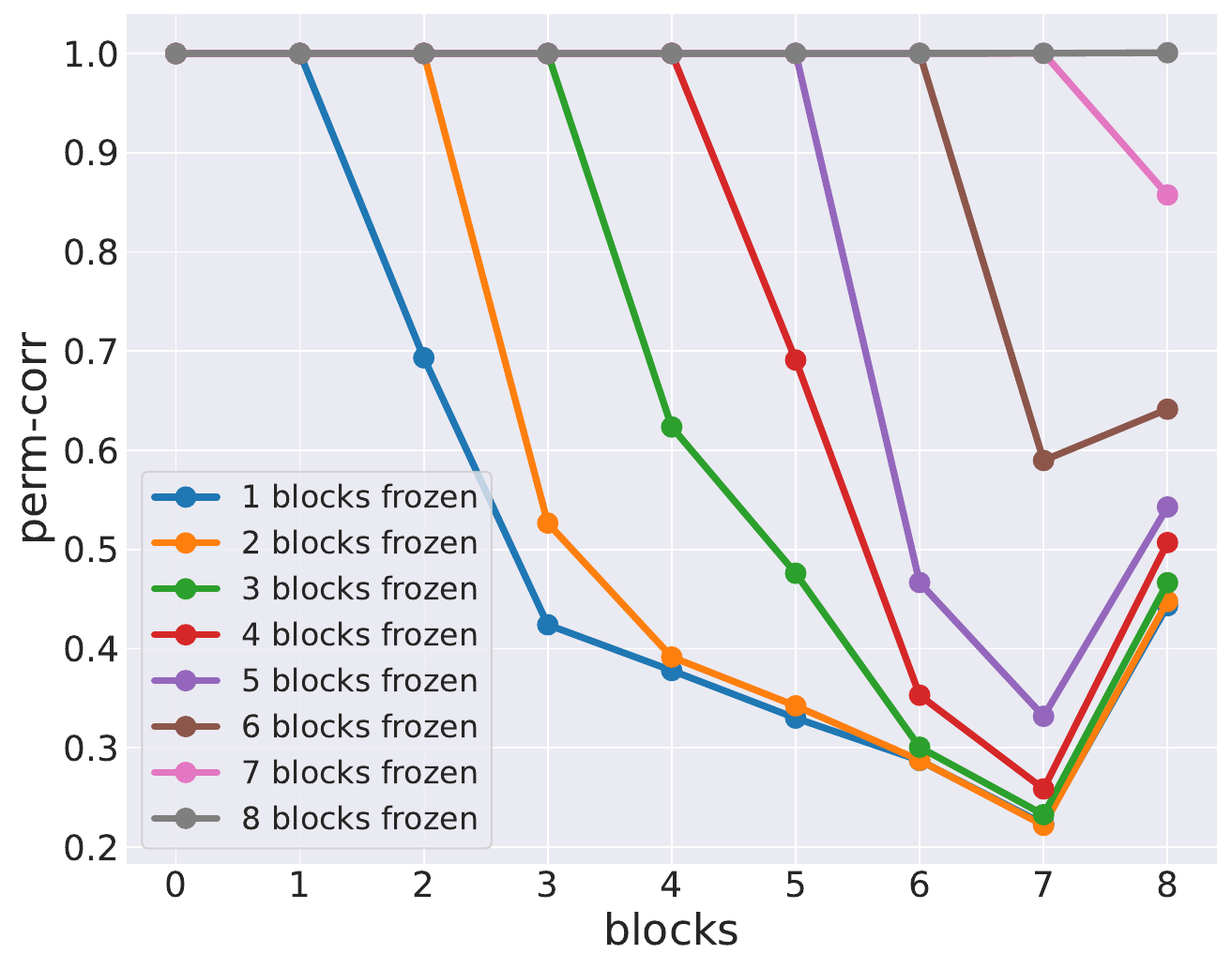}} \\
    {\includegraphics[width=0.31\textwidth]{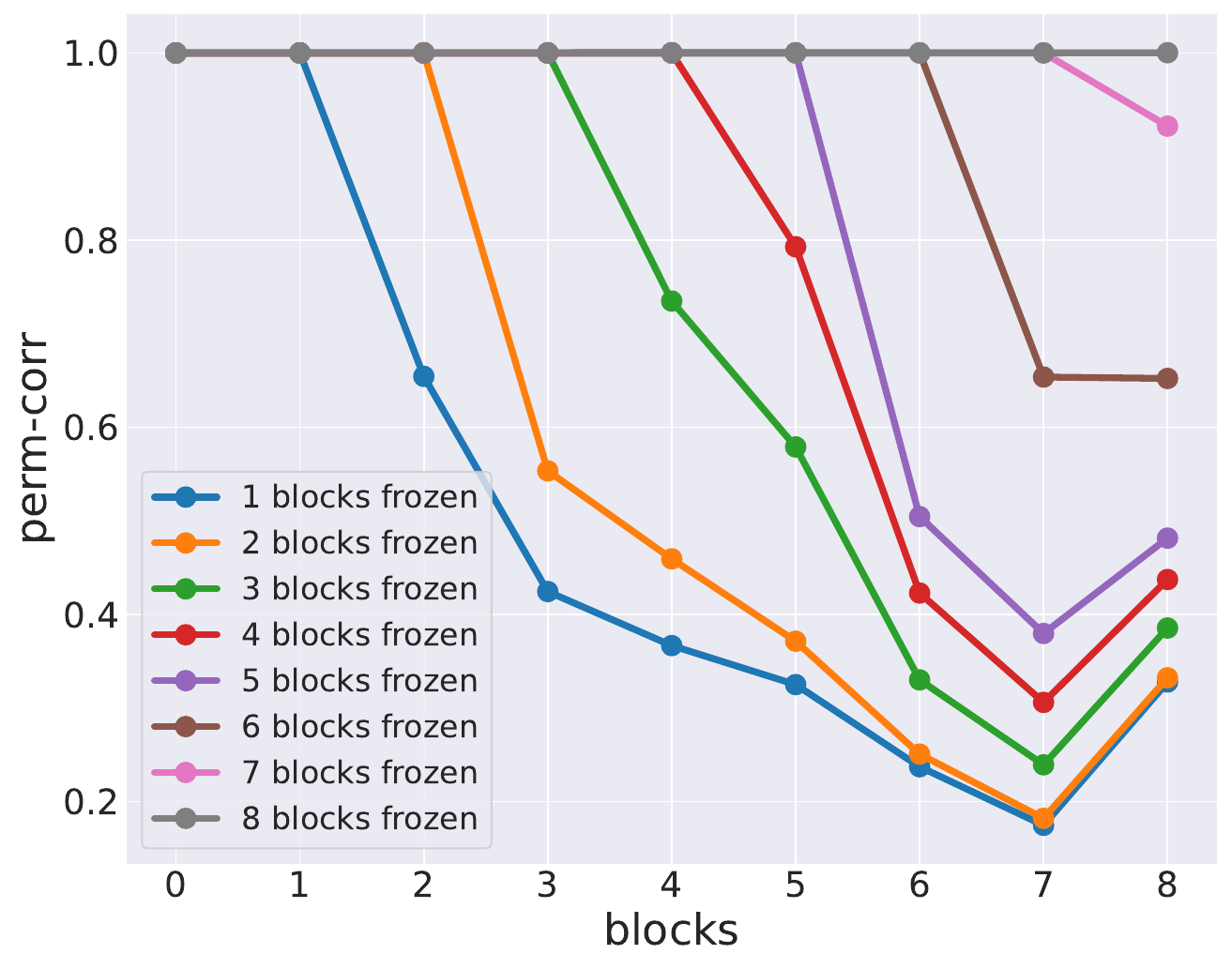}} \\
   {\includegraphics[width=0.31\textwidth]{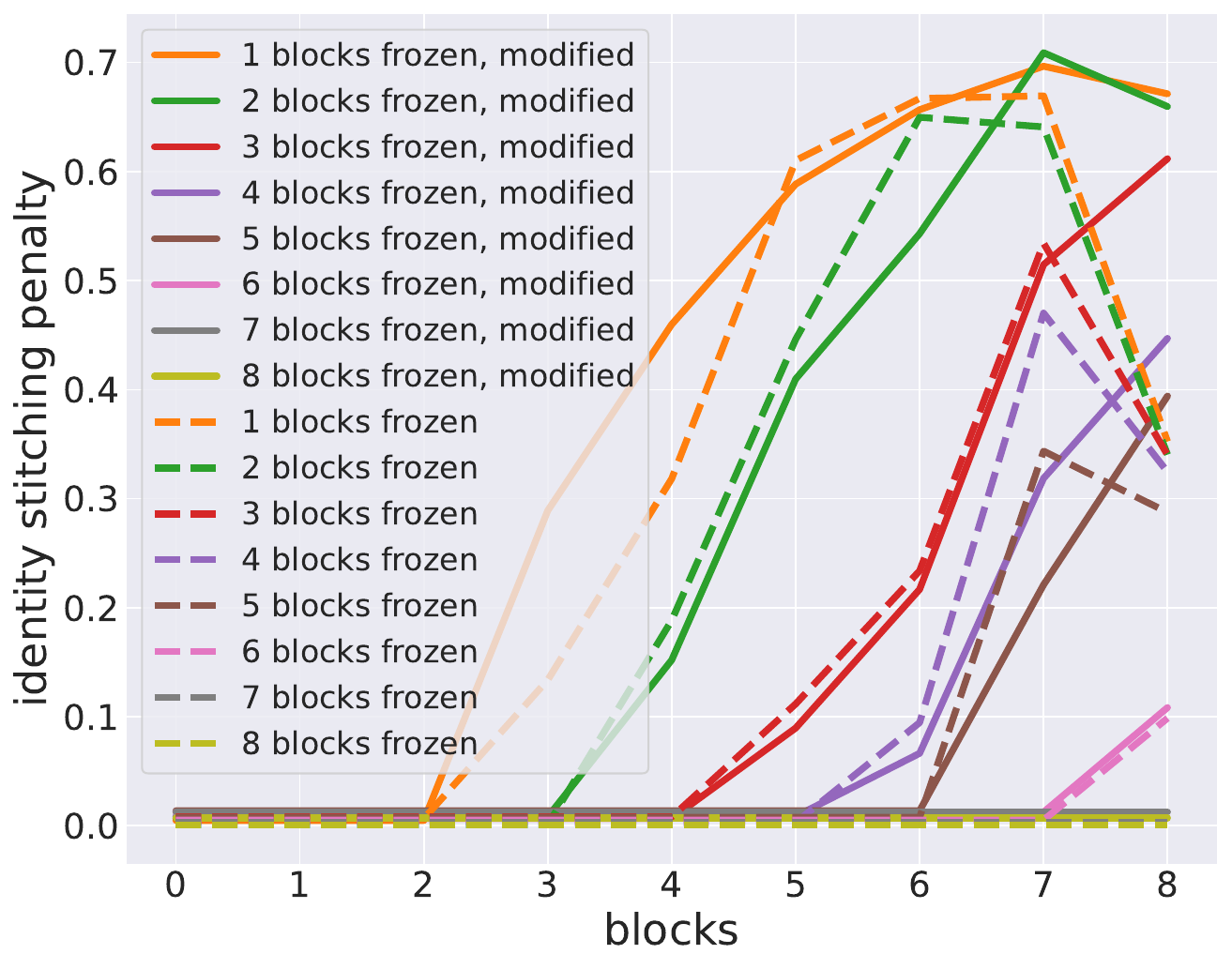}}
   \end{multicols}
    \caption{Measuring basis convergence for a normal ResNet-20 and a modified ResNet-20 when freezing $l$ early layers. The modified ResNet-20 is modified to have a fully linear residual stream (i.e., does not have a nonlinearity after the identity connection). (left) perm-corr for ResNet-20 models trained on CIFAR-10 (middle) perm-corr for ResNet-20 models trained on CIFAR-10 (right) identity stitching penalties for ResNet-20 and modified ResNet-20 models for CIFAR-10}
    \label{fig:res-stream-modified}
\end{figure*}

Here, we examine the phenonemon of convergent bases across 1) different model widths and 2) different residual stream structures. One plausible candidate for the converging basis phenomenon displayed in Figure \ref{fig:res-stream-convergence} is the structure of the residual stream. We modify a ResNet-20 so that there is no longer a non-linearity after skip-connections. For the modified ResNet-20, the residual stream is now completely `linear', where all layers exclusively perform linear operations on the residual stream (i.e., there are still nonlinearities within residual blocks, but no nonlinear operations on the residual stream). Figure \ref{fig:res-stream-modified} compares the perm-corr and identity stitching between a normal and modified ResNet-20. The results are largely comparable, suggesting that the convergent basis phenomonenon is not caused by the presence of a linear residual stream.

\begin{figure*}[!h]
    \centering
    \begin{multicols}{2}
    {\includegraphics[width=0.48\textwidth]{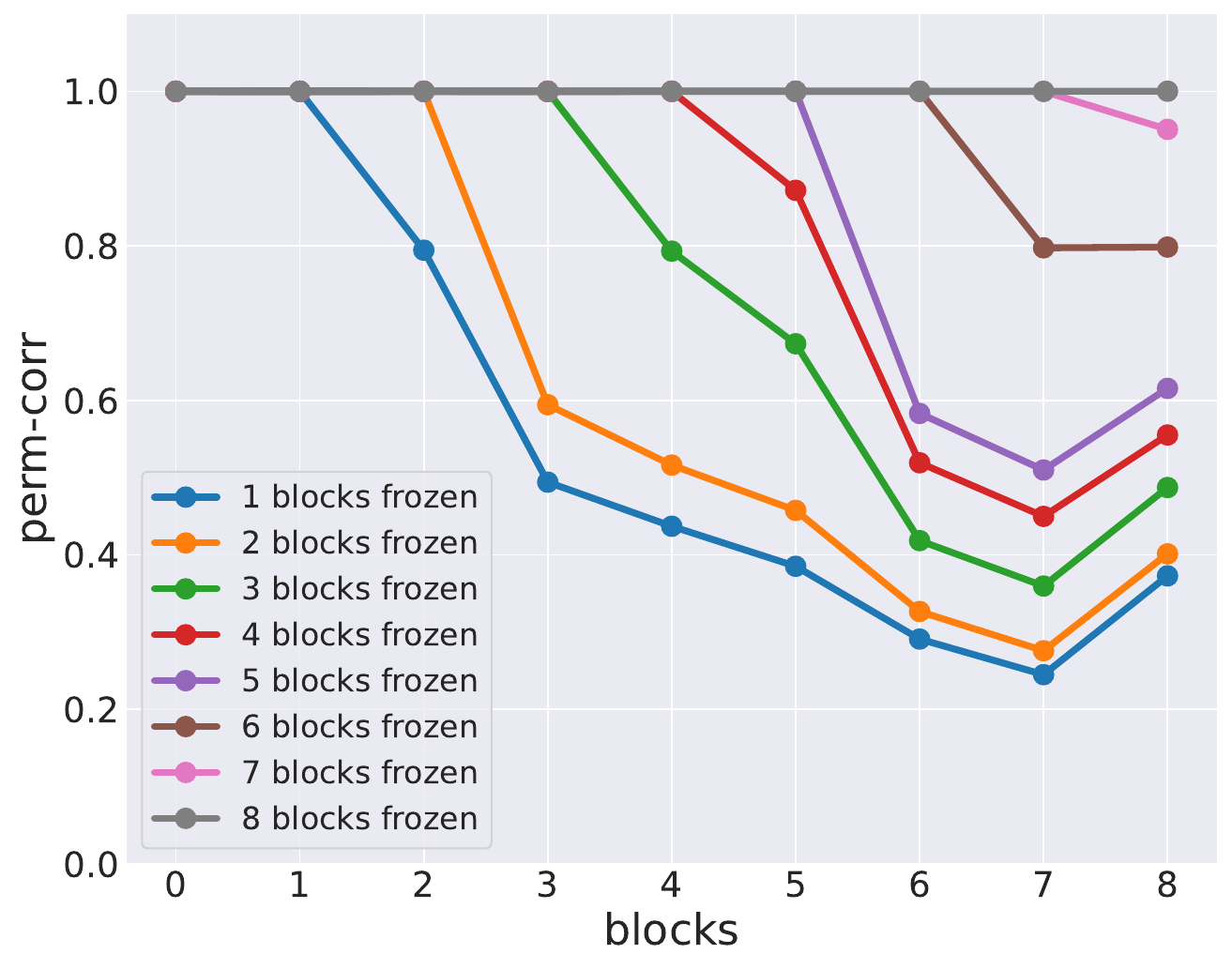}} \\
    {\includegraphics[width=0.48\textwidth]{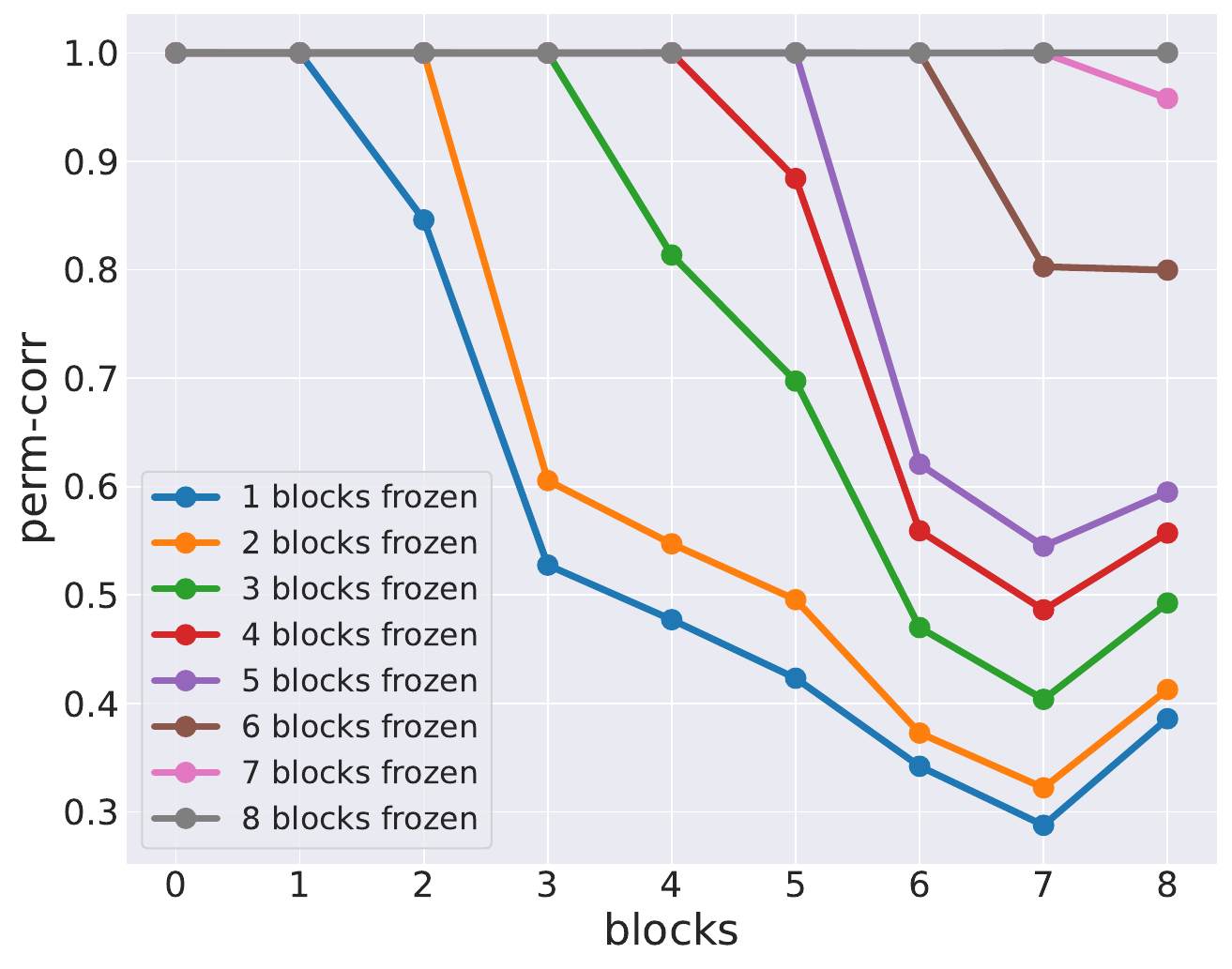}} \\
   \end{multicols}
\caption{Measuring perm-corr for wide ResNet-20s when freezing $l$ early layers (note different y-axis scales). (left) perm-corr for 4x-width ResNet-20 models trained on CIFAR-10 (right) 8x-width ResNet-20 models trained on CIFAR-10.}
    \label{fig:res-stream-convergence-perm-width}
\end{figure*}

\begin{figure*}[!h]
    \centering
    \begin{multicols}{2}
    {\includegraphics[width=0.45\textwidth]{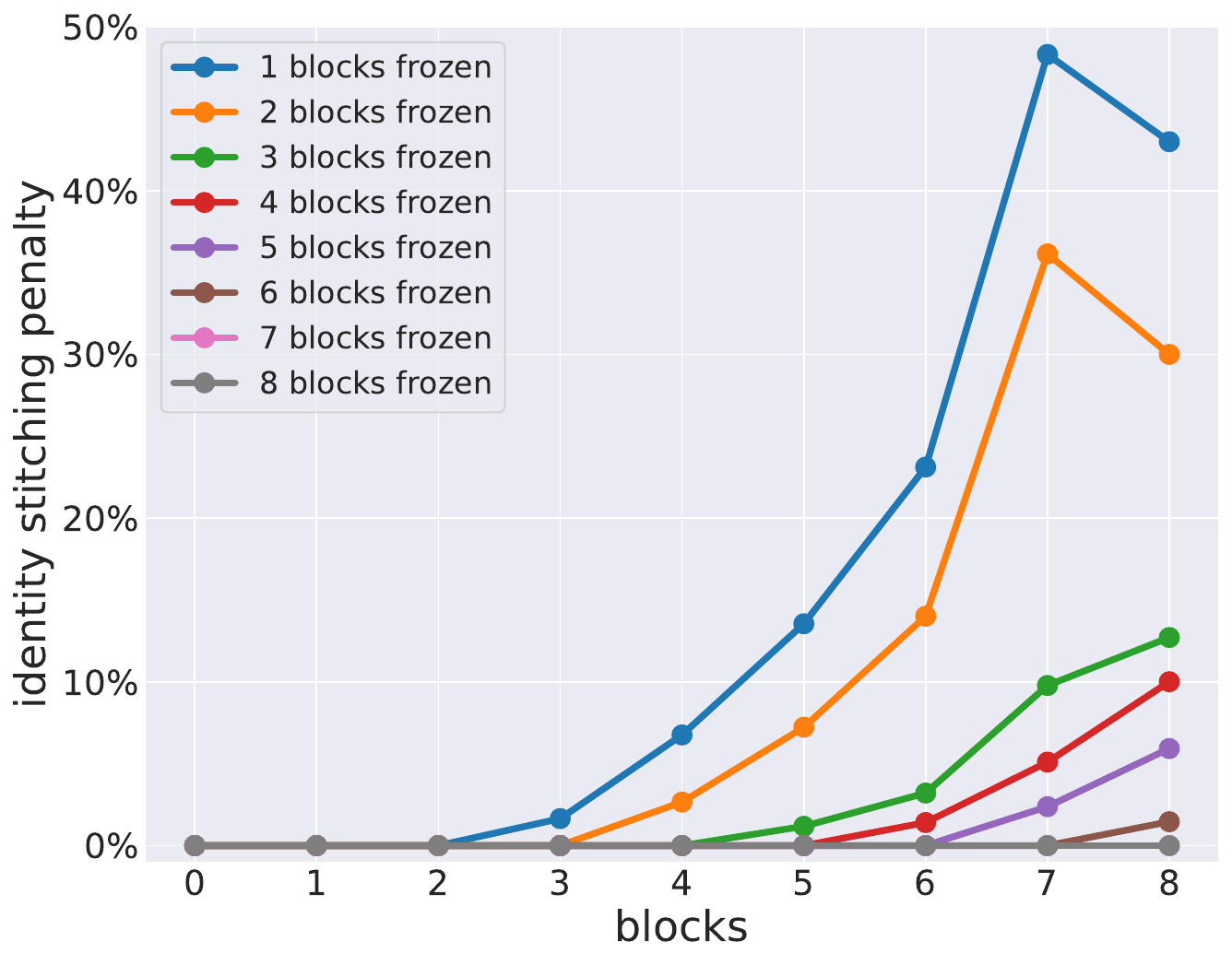}} \\ 
    {\includegraphics[width=0.45\textwidth]{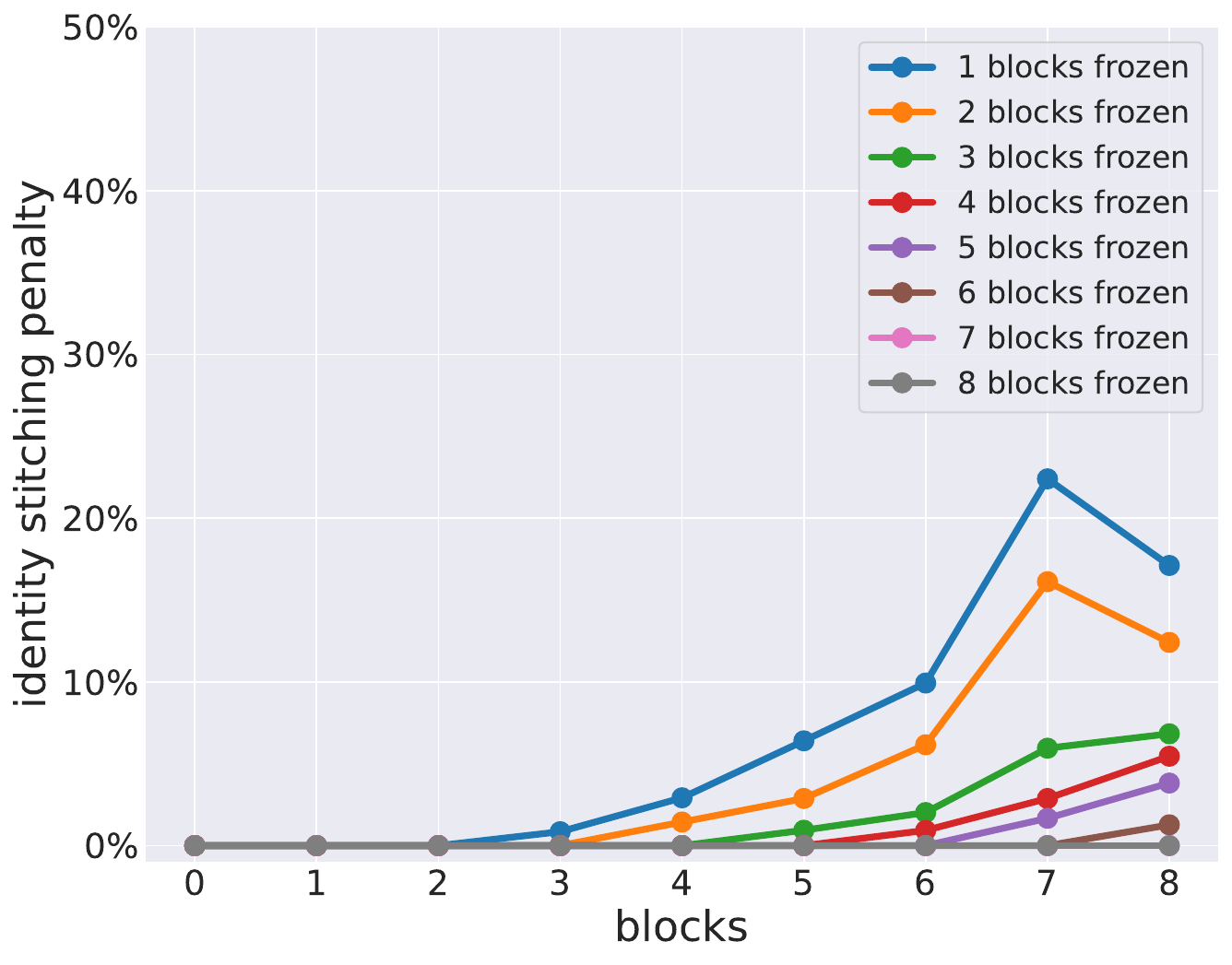}} \\
   \end{multicols}
    \caption{Measuring identity stitching for wide ResNet-20s when freezing $l$ early layers. (left) identity stitching penalty for 4x-width ResNet-20 models trained on CIFAR-10 (right) identity stitching penalty for 8x-width ResNet-20 models trained on CIFAR-10.}
    \label{fig:res-stream-convergence-stitching}
\end{figure*}

Next, we examine the effect of width. Figure \ref{fig:res-stream-convergence-perm-width} measures the perm-corr for a 4x-width and 8x-width ResNet-20, and Figure \ref{fig:res-stream-convergence-stitching} measures the identity stitching penalty for a 4x-width and 8x-width ResNet-20. We find that width can explain some of the identity stitching success, however there is little difference between the respective networks for perm-corr in Figure \ref{fig:res-stream-convergence-perm-width}.

\end{document}